\documentclass{article}

\usepackage{arxiv}

\usepackage[utf8]{inputenc}    
\usepackage[T1]{fontenc}       
\usepackage{textcomp}          
\usepackage{amsmath,amssymb,amsfonts} 
\usepackage{nicefrac}          
\usepackage{microtype}         
\usepackage{graphicx}          
\usepackage{booktabs}          
\usepackage{adjustbox}         
\usepackage{caption}           
\usepackage{xcolor}            
\usepackage{algorithm}
\usepackage{algpseudocode}
\usepackage[numbers]{natbib}   
\usepackage{url}
\usepackage{hyperref}          

\title{Optimizing ARDL Models for Retail Sales Forecasting and Fair Pricing}

\author{
 Sujay Uday Rittikar \\
  Department of Applied Computer Science \\
  The University of Winnipeg \\
  Winnipeg, MB, Canada \\
  \texttt{rittikar-s@webmail.uwinnipeg.ca} \\
  \texttt{suj00rit20@gmail.com} \\
}

\begin{document}
\maketitle

\begin{abstract}
Pricing food products to balance profitability with consumer welfare is a central challenge for retailers. Dynamic pricing is widely used to maximize revenue, yet most pricing models optimize business objectives while overlooking consumer fairness. This paper studies the risk of consumer exploitation under dynamic food pricing in Canada and proposes a methodology that embeds fairness constraints directly into retail sales forecasting. We model total retail trade sales with a log--log Autoregressive Distributed Lag (ARDL) specification, in which the coefficient on a product price is a sales elasticity, and pose the pricing problem as maximizing forecast sales subject to price bounds anchored to the Consumer Price Index (CPI). We solve this problem with both Linear Programming (LP) and Simulated Annealing (SA), under single-product and multi-product configurations. A key finding is that the fitted nominal elasticities are positive. As a result, an unconstrained sales-maximizer would push every price to its upper bound, and the CPI ceiling is the safeguard that prevents this. Simulated Annealing instead settles on conservative, interior prices that lower consumer cost while still meeting the sales target. We benchmark forecast accuracy against naive, seasonal-naive, ARIMA, and SARIMA baselines, and a CPI-deflated re-specification shows that the positive nominal elasticities are largely an inflation-driven artifact. The result is a transparent, fairness-aware pricing framework.
\end{abstract}

\keywords{Retail Sales Forecasting \and Autoregressive Distributed Lag \and Consumer Fairness \and Price Elasticity \and Linear Programming \and Simulated Annealing \and Metaheuristics}

\section{Introduction}
Retail food prices are adjusted based on factors affecting a business's revenue and sales, with dynamic pricing being a common strategy to optimize profits. This involves varying prices over time or across customer segments to maximize their revenue. While effective, dynamic pricing raises concerns about fairness \cite{Levin2010}. Many consumers are unaware of complex pricing strategies, leading to perceptions of unfairness \cite{Deane2017,Xia2004}. Balancing price optimization with fairness is essential. However, price optimization often relies on behavioral biases rather than genuine differences in consumer valuations \cite{Kallus2021}, which can lead to unethical practices when businesses prioritize profit over fairness \cite{Seele2021}. Personalized pricing can increase efficiency but may trigger backlash if price differences are linked to protected characteristics like gender or race, making it difficult to distinguish between legitimate preferences and unfair discrimination.

Retailers often adjust prices based on historical trends to boost sales and enhance profitability, which highlights the challenge of time-series forecasting. This process involves predicting retail sales while accounting for price fluctuations over time \cite{Nijs2007}. With the advancement of machine learning, various statistical time-series forecasting models have emerged, one of the most prominent being the Autoregressive Distributed Lag (ARDL) model. ARDL incorporates exogenous variables to make predictions, making it particularly effective for long-term forecasting \cite{Kripfganz2023}. The model has been widely applied in addressing economic problems such as analyzing trade openness and economic growth \cite{Kong2021}. To optimize the performance of machine learning models and align their sales forecasts closely with actual retail sales, various strategies are employed. A common one is to combine multiple forecasting models: the large-scale M5 retail forecasting competition found that combinations and ensembles of models, particularly gradient-boosted trees, ranked among the strongest performers on retail sales data \cite{Makridakis2022M5}, and stacking has been used to leverage the strengths of different predictive techniques \cite{Pavlyshenko2019}. A second strategy embeds optimization methods such as linear programming, which introduce constraints that fine-tune predictions and align them with business and fairness objectives \cite{Ganguly2024}. More recently, high-capacity models such as time-series transformers have been evaluated specifically for retail demand forecasting \cite{Oliveira2024}. In parallel, researchers have increasingly applied metaheuristics to optimize time-series forecasts, even in complex, multi-variable scenarios \cite{Zito2023}, keeping forecasts realistic and consistent with market constraints.

Retail sales forecasting is already a complex task, and ensuring consumer fairness adds a further layer of responsibility. It requires optimizing prices to prevent exploitation, a goal that can be anchored to metrics such as the Canadian Consumer Price Index (CPI) \cite{Fradella2022}. The CPI tracks monthly price changes across the Canadian retail market, reflecting inflation trends, and is categorized by product groups such as food, health, transportation, and more. By adjusting prices in line with CPI trends, businesses can align their strategies with inflation, ensuring fairness and transparency. CPI-adjusted prices also help meet consumer expectations, fostering trust and equity in the market. This paper presents a new method for forecasting retail sales, aiming to maximize sales while keeping prices close to CPI-adjusted retail values. We achieve this by optimizing the ARDL model, adding extra constraints, and addressing the problem in two different scenarios: one with a single exogenous feature and another with multiple exogenous features. In this context, an exogenous feature refers to the retail prices of a food product. To tackle this complex multivariable time-series forecasting problem, we use Linear Programming and the Simulated Annealing Metaheuristic Algorithm to model the forecasting process while optimizing the objectives.

\section{Related Work}

Machine learning models have become widely used for time-series forecasting, including sales forecasting. Pavlyshenko \cite{Pavlyshenko2019} combined the outputs of several models through stacking, improving accuracy on both validation and out-of-sample data, particularly for new products. Using the Rossmann Store Sales data from Kaggle \cite{FlorianKnauer2015}, that study argues sales forecasting is better framed as a regression problem, since regression-oriented models outperformed pure time-series models. This conclusion is drawn mainly from univariate settings, however, and may not carry over to multivariate ones. The same work notes that relying on historical data alone to capture seasonality, together with the absence of exogenous variables, produced noisy forecasts, underscoring the value of external factors. Later work applied a broader range of regression models. Ganguly et al. \cite{Ganguly2024} pointed to the limitations of linear regression (LR) on retail data with seasonality and external effects, and compared Random Forest (RF), Gradient Boosting (GB), Support Vector Regression (SVR), and XGBoost (XGB) on the Favorita Store Sales dataset \cite{AlexisCook2021}. Hyperparameter tuning raised the RF R-squared from 0.915 to 0.945, above LR's 0.531, with the tuned RF slightly ahead of GB, XGB, and SVR, which performed comparably. Consistent with Pavlyshenko \cite{Pavlyshenko2019}, they identified data noise and missing exogenous variables as the main obstacles to long-term accuracy. Recent retail studies reinforce this point: Hewage et al. \cite{Hewage2025} quantify the effect of sales promotions across the demand life cycle, showing how strongly such external signals shape forecast quality, while explainable gradient-boosting methods such as the LightGBM-SHAP framework of Mishra et al. \cite{Mishra2026} pair competitive accuracy with interpretable demand drivers. A separate concern is that models excelling on validation data can behave inconsistently once deployed, owing to distribution shift, outliers, and the general messiness of real settings. Metaheuristic search offers one remedy by constructing constrained artificial environments in which forecasts are steered toward realistic, if suboptimal, behaviour. Zito et al. \cite{Zito2023} proposed a local-search metaheuristic that optimizes such environments for a predictor, embedding constraints and observed conditions so the model better mirrors real-world dynamics. Metaheuristics suit non-linear, high-dimensional data, and recent hybrids extend this idea: Zhao et al. \cite{Zhao2026} apply a modified metaheuristic to tune a deep time-series model, illustrating how metaheuristic optimization and machine learning can combine for scalable forecasting in domains such as retail sales and energy demand.

Linear programming offers another route to optimizing time-series forecasts. Selvam et al. \cite{Selvam2024} proposed a Linear Programming-based Bi-Objective Forecasting Algorithm (BOFA) that fits Ordinary Least Squares (OLS) and neural-network regression models under two simultaneous objectives, minimizing the Mean Absolute Error (MAE) and the Maximum Absolute Error (MaxAE). Balancing the two curbs both over- and under-forecasting, and the method reported gains over standard baselines such as ARIMA and Holt-Winters across short- and long-term horizons.
A recurring theme across these studies is that omitting exogenous variables yields noisy, unreliable forecasts. The Autoregressive Distributed Lag (ARDL) model \cite{Kripfganz2023} addresses this directly by including lagged values of both the dependent variable and exogenous regressors, with the lag structure chosen by criteria such as the Akaike or Bayesian Information Criterion (AIC/BIC). The framework also supports the bounds-testing procedure of Pesaran et al. \cite{Pesaran2001}, which detects long-run cointegrating relationships among variables that may be integrated of different orders, making ARDL well suited to macroeconomic series that are not uniformly stationary in levels. As a single-equation model, it captures the dynamic interplay between exogenous and target variables and improves accuracy on data shaped by external factors. As one applied example, ARDL has been used on the grain industry across EU countries to study the effects of climate, technology, and marketing \cite{Mandych2024}, demonstrating its ability to handle non-stationary data and dynamic interrelationships.

Beyond classical econometric models, deep learning has produced a family of high-capacity forecasters, including Prophet \cite{Taylor2018}, DeepAR \cite{Salinas2020}, and the Temporal Fusion Transformer \cite{Lim2021}, which capture complex non-linear and multi-horizon dynamics. These models are, however, data-intensive and largely opaque, whereas our objective requires an interpretable, low-parameter relationship between an individual product price and aggregate sales that can be embedded directly into an optimization program. We therefore adopt the ARDL model, trading raw predictive capacity for the transparency and economic interpretability, namely, price elasticities needed to reason about fairness. On the fairness side, perceived price (un)fairness has long been studied in marketing \cite{Xia2004}, and recent work has mapped the ethical concerns raised by dynamic and personalized pricing \cite{Seele2021,Kallus2021}. This literature motivates the use of external, transparent reference points for pricing. Our use of CPI-anchored price bounds operationalizes such a reference point, connecting the fairness literature to a concrete forecasting-and-optimization pipeline.

\section{Problem Formulation and Methodology}
The problem is approached from the perspectives of both retailers and consumers, aiming to forecast maximum retail sales for food products while ensuring prices do not exceed CPI-adjusted levels. These prices act as a benchmark to prevent exploitation through dynamic pricing and ensure fairness. Our methodology employs Autoregressive Distributed Lag (ARDL) models, capturing the dynamic relationships between sales and prices over time. These models are optimized using Linear Programming (LP) and Simulated Annealing metaheuristics for reliable forecasts. The approach balances maximizing sales with fair pricing, aligning with economic goals and consumer welfare.

\subsection{Dataset Overview}\label{AA}
Our dataset is constructed using monthly data from Statistics Canada, including the Consumer Price Index (CPI) \cite{StatisticsCanada2018}, retail prices by food product \cite{StatisticsCanada2020}, and retail trade sales for all food products \cite{StatisticsCanada2023}. The data spans from January 2017 to August 2024 and covers all provinces of Canada. However, for this study, we analyze the aggregated data for the entire country without considering provincial segregation.

The Canadian Consumer Price Index (CPI) is a widely used standard for measuring the average price changes of products over time. Since the CPI is categorized by product type, we specifically focus on the Food Product CPI. The base CPI is indexed at 100, with the base year being 2002. For our analysis, the CPI value for January 2017, which is 141.5, is considered as the base for our dataset. The monthly retail price data provides the average retail prices, measured in Canadian dollars, for 106 food products across Canada for each month. Similarly, the monthly retail sales data represents the total retail sales, also measured in Canadian dollars, for all food products nationwide. By examining the relationship between retail prices of individual food products and total retail sales, we aim to understand how product-level pricing influences macroeconomic retail sales trends across the country.

From the 106 food products, we focus on five key items: apples, eggs, milk, white bread, and roasted or ground coffee. Table \ref{tab:stat_summary} provides the statistical summary for these selected columns, including their minimum, maximum, mean, and standard deviation (SD) values.

\begin{table}[h!]
\caption{Statistical Summary of Data}
\centering
\begin{adjustbox}{width=0.7\linewidth}
\begin{tabular}{lrrrr}
\toprule
\textbf{Feature Name} & \textbf{Min} & \textbf{Max} & \textbf{Mean} & \textbf{SD} \\
\midrule
CPI & 141.5 & 190.5 & 160.914 & 16.217 \\
Retail Sales (\$) & 3.76e+10 & 6.69e+10 & 5.82e+10 & 6.40e+09 \\
Apples (per kg) & 3.33 & 6.09 & 4.733 & 0.701 \\
Eggs (1 dozen) & 3.01 & 4.74 & 3.825 & 0.530 \\
Milk (2 litres) & 4.01 & 5.29 & 4.489 & 0.403 \\
White Bread (675 g) & 2.73 & 3.74 & 3.100 & 0.296 \\
Roasted/Ground Coffee (340 g) & 4.59 & 7.09 & 5.620 & 0.622 \\
\bottomrule
\end{tabular}
\end{adjustbox}
\label{tab:stat_summary}
\end{table}

\subsection{Data Partitioning and Preliminary Analysis}
We split the dataset chronologically into training (80\%) and test (20\%) sets. Training spans January 2017 to February 2023, and the test period covers the 18 months from March 2023 to August 2024, placing this study in the long-term forecasting regime. The analyses below characterize the series and, in doing so, motivate three later modeling choices: the plausibility bounds on sales, the use of an ARDL specification, and the CPI-anchored price ceiling.

\begin{figure}[ht!]
    \centering
    \includegraphics[width=\linewidth]{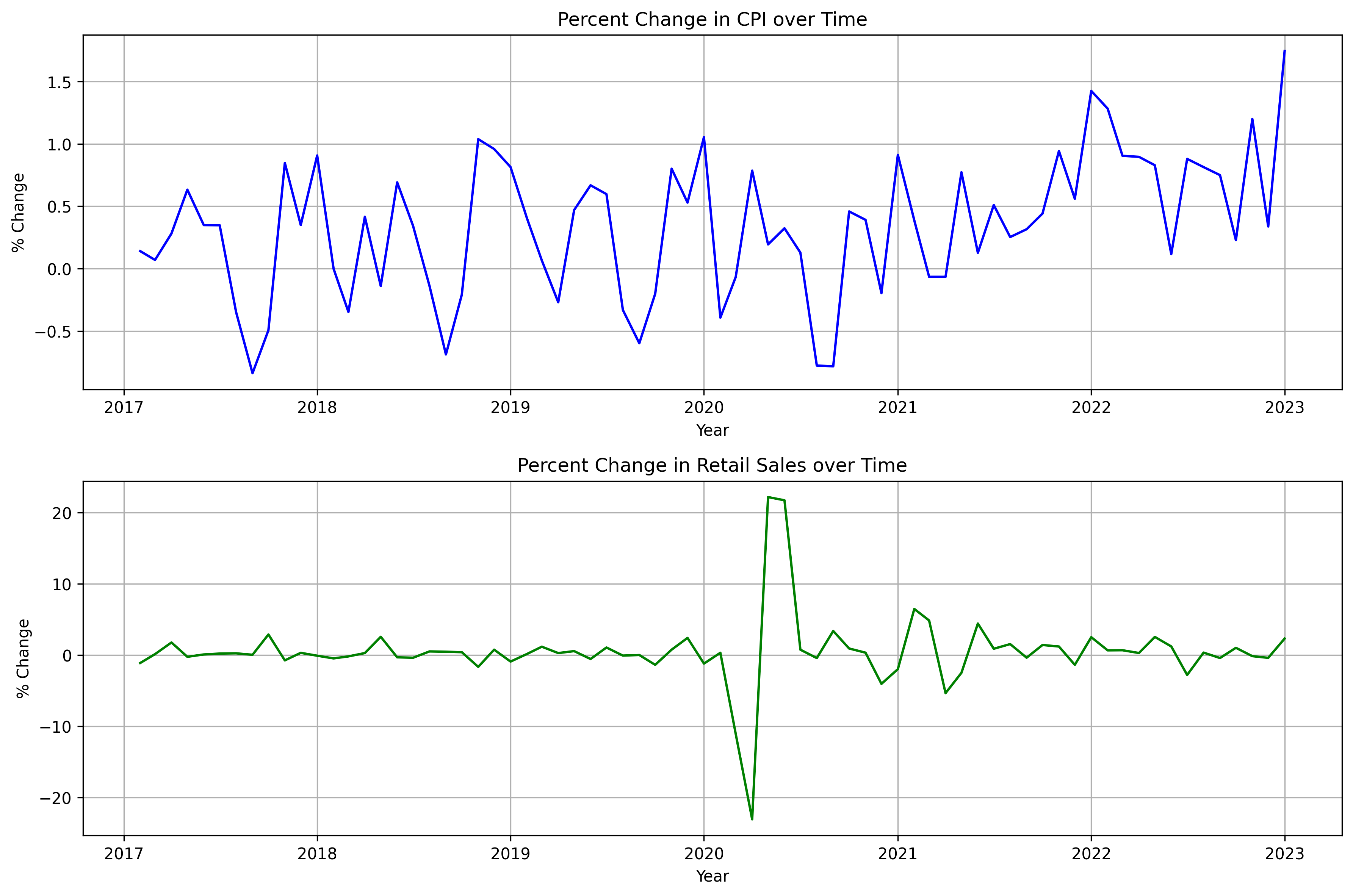}
    \caption{Percent change in CPI and Retail Sales over Time.}
    \label{fig:percent_change}
\end{figure}

We first examine the month-over-month percentage changes in CPI and retail sales to gauge how far sales realistically move. As shown in Fig. \ref{fig:percent_change}, retail sales range from a maximum change of 22.19\% to a minimum of -23.09\%, but these extremes each occur only twice and appear driven by external shocks; sales otherwise fluctuate within roughly -23\% to 22\%. The CPI is far steadier, varying between 1.74\% and -0.83\%. This observed range directly informs the admissible sales interval imposed in the optimization program of Section~\ref{sec:methodology}.

These growth rates are close to stationary aside from a few outliers, whereas the series in \emph{levels} continue to trend upward. An Augmented Dickey--Fuller test in Section~\ref{sec:methodology} confirms that total retail sales contain a unit root, which the ARDL specification is designed to accommodate.

Finally, we assess whether retail prices track the CPI, since this is what licenses using CPI-adjusted values as a fairness benchmark. Table \ref{tab:cpi_retail_correlation} reports Pearson correlation coefficients between each food product's price and the CPI. All items exceed 0.94 except apples (0.77); the weaker apple correlation suggests product-specific factors beyond general inflation. The consistently strong coupling supports anchoring prices to the CPI to model price dynamics, absorb inflation, and limit market distortion.

\begin{table}[ht!]
    \centering
    \caption{Pearson's Correlation Coefficient between CPI and Retail Prices}
    \begin{tabular}{|l|c|}
        \hline
        \textbf{Product} & \textbf{Pearson's Correlation Coefficient} \\
        \hline
        Apples (per kilogram) & 0.77 \\
        Eggs (per dozen) & 0.94 \\
        Milk (2 liters) & 0.98 \\
        White bread (675 grams) & 0.95 \\
        Roasted/Ground coffee (340 grams) & 0.95 \\
        \hline
    \end{tabular}
    \label{tab:cpi_retail_correlation}
\end{table}

\subsection{Methodology}\label{sec:methodology}
Because retail sales, prices, and the CPI span very different magnitudes and exhibit multiplicative (percentage) dynamics, we model all variables on the natural-logarithm scale. A log--log ARDL specification has the convenient property that the coefficient on a log price is a \emph{sales elasticity}: it measures the approximate percentage change in retail sales associated with a one-percent change in that price, and forecasts are exponentiated back to the original dollar scale for evaluation. To confirm that this specification is appropriate, we performed the Augmented Dickey--Fuller (ADF) test on log retail trade sales using the \texttt{statsmodels} Python library \cite{Seabold2010}, with the lag length selected by the Akaike Information Criterion (AIC). On the training sample the test statistic is $-1.25$ ($p = 0.65$), so the level series is non-stationary and contains a unit root; after first differencing the statistic falls to $-9.37$ ($p < 0.001$), confirming that the series is integrated of order one, $I(1)$. We therefore adopt an ARDL model with two autoregressive lags (again selected by AIC) within the error-correction-consistent ARDL framework \cite{Kripfganz2023,Pesaran2001}, which accommodates such $I(1)$ dynamics without pre-differencing the exogenous regressors.

We use a dynamic configuration, meaning the model feeds its own previously predicted values back as inputs for multi-step-ahead prediction, as required for long-term forecasting. The fitted single-product model is written as the log--log linear equation:
\begin{equation}
\ln y_t = \alpha + \beta_1 \ln y_{t-1} + \beta_2 \ln y_{t-2} + \gamma \ln x_t + \epsilon_t
\end{equation}
where:
\begin{itemize}
    \item \(y_t\): the retail trade sales at time \(t\) (modelled as \(\ln y_t\)),
    \item \(\ln y_{t-1}, \ln y_{t-2}\): the lagged log values of sales,
    \item \(x_t\): the retail price of the food product at time \(t\),
    \item \(\alpha\): the intercept term,
    \item \(\beta_1, \beta_2\): the autoregressive coefficients on lagged log sales,
    \item \(\gamma\): the price elasticity of sales (percentage change in sales per one-percent change in price),
    \item \(\epsilon_t\): the error term.
\end{itemize}

In the multi-product configuration, the single term \(\gamma \ln x_t\) is replaced by a sum \(\sum_{j} \gamma_j \ln x_{j,t}\) over the selected products, so that each \(\gamma_j\) is the partial elasticity of aggregate sales with respect to product \(j\).

Using the ARDL model fitted to the training data, we define a single-period pricing optimization problem. Holding the recent log-sales history $\ln y_{t-1}, \ln y_{t-2}$ and the estimated coefficients fixed, we treat the log prices as decision variables and maximize predicted sales subject to fairness (price) and plausibility (sales) bounds:
\begin{align}
\max_{\{\ln x_{j}\}} \quad & \ln \hat{y} = \alpha + \beta_1 \ln y_{t-1} + \beta_2 \ln y_{t-2} + \sum_j \gamma_j \ln x_j \notag\\
\text{s.t.} \quad & \ln p_j^{\min} \le \ln x_j \le \ln p_j^{\max}, \quad \forall j, \label{eq:program}\\
& \ln s^{\min} \le \ln \hat{y} \le \ln s^{\max}, \notag
\end{align}
where $p_j^{\max}$ is the CPI-adjusted price of product $j$ (the fairness ceiling), $p_j^{\min} = 0.65\,p_j^{\max}$, and $[s^{\min}, s^{\max}]$ bounds the admissible sales. This model is stated procedurally in Algorithm \ref{alg:opt_algorithm}.

Because the objective in \eqref{eq:program} is linear in the decision variables and the feasible region is a box intersected with the sales bounds, the problem is a linear program whose optimum lies at a vertex of the feasible region. In particular, the sales-maximizing price for product $j$ is driven to its upper bound $p_j^{\max}$ when the fitted elasticity $\gamma_j > 0$ and to its lower bound when $\gamma_j < 0$; the solution is therefore a deterministic corner point. We solve this exact program with Linear Programming (LP) to obtain the tight-bound recommendation, which characterizes the extreme of what a pure sales-maximizer would charge.

Corner solutions, however, are brittle from a fairness standpoint: whenever $\gamma_j > 0$ which, as Section~\ref{sec:results} shows, holds for every product in the nominal single-product models, the LP recommends the most expensive admissible price. We therefore also solve the problem with the Simulated Annealing (SA) metaheuristic \cite{Kirkpatrick1983}, a stochastic search that does not commit to a vertex. Starting from a random feasible price and perturbing it under a cooling schedule while respecting the sales cap, SA is a finite stochastic search that need not reach the exact vertex: a fully converged run coincides with the LP corner, whereas a run with a conservative stopping criterion terminates at feasible interior prices. In this setting SA is thus not a device for escaping local optima, the LP optimum is global and known, but a way to trade a small amount of predicted sales for markedly lower, more consumer-favourable prices.

For the Simulated Annealing Metaheuristic approach, we employ a temperature-based strategy. If the new solution for retail price(s) is optimal, it is always accepted. Otherwise, the acceptance depends on the probability given by the following formula:
\begin{equation}
P = e^{-\frac{\Delta E}{T}}
\end{equation}
where:
\begin{itemize}
    \item \( \Delta E \) is the change in the objective function value.
    \item \( T \) is the current temperature.
    \item \( P \) is the probability of accepting a worse solution.
\end{itemize}

As the algorithm progresses, the temperature is gradually reduced following a cooling schedule:
\begin{equation}
T_{\text{new}} = \alpha T_{\text{old}}
\end{equation}
where \( \alpha \) is a constant cooling factor. This reduction in temperature allows the algorithm to focus more on exploiting the best solutions found so far, as it becomes less likely to accept worse solutions.

\section{Experiment and Evaluation}
This section describes how the models were implemented, the optimization settings used, and the metrics adopted to evaluate the forecasts.

\begin{algorithm}[h]
\caption{Optimization Model for Retail Sales Forecasting}
\label{alg:opt_algorithm}
\begin{algorithmic}[1]
\Require Lagged log-sales $\ln y_{t-1}$, $\ln y_{t-2}$; number of products $n$ (default $n{=}1$);
  fitted ARDL coefficients $\alpha, \beta_1, \beta_2, \gamma_1, \dots, \gamma_n$;
  price bounds $p_i^{\min}, p_i^{\max}$; sales bounds $s^{\min}, s^{\max}$
\Ensure Log-prices $\ln x_i$ and predicted log-sales $\ln \hat{y}$
\Statex
\State \textbf{maximize} \quad $\ln \hat{y}$
\Statex
\State \textbf{subject to}
\State \quad $\ln \hat{y} = \alpha + \beta_1 \ln y_{t-1} + \beta_2 \ln y_{t-2} + \sum_{i=1}^{n} \gamma_i \ln x_i$ \Comment{ARDL prediction}
\State \quad $\ln p_i^{\min} \le \ln x_i \le \ln p_i^{\max}, \quad i=1,\dots,n$ \Comment{price bounds}
\State \quad $\ln s^{\min} \le \ln \hat{y} \le \ln s^{\max}$ \Comment{sales bounds}
\end{algorithmic}
\end{algorithm}

\subsection{Implementation and Setup}
We developed the Linear Programming model in Python, utilizing the amplpy library\protect\footnotemark[1] to integrate with AMPL models. Additionally, we implemented the Simulated Annealing approach in Python, leveraging the numpy library \cite{Harris2020} to generate random numbers for probability estimation and decision-making. The optimization model was implemented as outlined in Algorithm \ref{alg:opt_algorithm}. The code was executed on a system running Windows 11, equipped with an Intel i5-12500H CPU @ 2.5GHz and 16 GB of RAM.
\footnotetext[1]{amplpy is a Python API for integrating AMPL with Python. Documentation: \url{https://ampl.com/tryampl/python-api/}.}

The problem was solved using two configurations:
\begin{enumerate}
    \item \textbf{Single-product Multivariate Time Series Forecasting:}
    In this configuration, we consider only one food product as an exogenous variable to model the Retail Sales Time Series. Specifically, we create five ARDL models, each corresponding to one of the five food products selected for this study.
    \item \textbf{Multi-product Multivariate Time Series Forecasting:}
    In this configuration, we consider all five food products chosen for this study as exogenous variables to model the Retail Sales Time Series. As a result, only one ARDL model is created, incorporating all five food products.
\end{enumerate}

\subsection{Optimization and Metaheuristic Settings}
As detailed in Algorithm \ref{alg:opt_algorithm}, the optimization model incorporates minimum and maximum thresholds for Retail Prices and Retail Trade Sales. In our study, the minimum retail price threshold is set at 65\% of the CPI-adjusted prices, while the maximum retail price threshold corresponds to the CPI-adjusted prices.

To emphasize our goal of maximizing retail trade sales, we define the target as 15\% above the ARDL-predicted sales. Accordingly, the minimum retail trade sales threshold is set to the ARDL-predicted sales, and the maximum retail trade sales threshold is 15\% higher. The target maximum retail trade sales can be adjusted based on domain knowledge of the retail market. For the metaheuristic Simulated Annealing algorithm, we use the following configuration:
\begin{itemize}
    \item $T = 1000$: Initial temperature
    \item $\alpha = 0.95$: Cooling rate
    \item $max\_iters = 500$: Iterations per temperature
    \item $\epsilon = 1 \times 10^{-5}$: Convergence tolerance
\end{itemize}

\subsection{Evaluation Metrics}
To evaluate the forecast results, we use the following metrics:
\begin{enumerate}
    \item \textbf{Root Mean Squared Error (RMSE)}: 
    The RMSE measures the significance of errors by evaluating the deviation of the forecasts from the original values in the test dataset, giving more weight to larger errors. It is defined as:
    \begin{equation}
    \text{RMSE} = \sqrt{\frac{1}{n} \sum_{i=1}^n \left( y_i - \hat{y}_i \right)^2}
    \end{equation}

    \item \textbf{Mean Absolute Error (MAE)}: 
    The MAE assesses the average magnitude of the errors in the forecast, providing an overall measure of how close the forecasts are to the actual values. It is defined as:
    \begin{equation}
    \text{MAE} = \frac{1}{n} \sum_{i=1}^n \left| y_i - \hat{y}_i \right|
    \end{equation}

    \item \textbf{Mean Absolute Percentage Error (MAPE)}: 
    The MAPE score indicates the average percentage difference between the forecasted values and the original values, offering an easily interpretable metric for forecasting accuracy. It is defined as:
    \begin{equation}
    \text{MAPE} = \frac{1}{n} \sum_{i=1}^n \left| \frac{y_i - \hat{y}_i}{y_i} \right| \times 100
    \end{equation}

    \item \textbf{Percentage Change (\% Change)}: 
    The Percentage Change values represent the relative change between the forecasted value and the actual values, expressed as a percentage. It is calculated as:
    \begin{equation}
    \% \text{Change} = \frac{1}{n} \sum_{i=1}^n \frac{\hat{y}_i - y_i}{y_i} \times 100
    \end{equation}
    A positive value indicates an average increase in the forecasted values compared to the ARDL-forecasted value, while a negative value indicates an average decrease.
\end{enumerate}

The terms used to define the above metrics:
\begin{itemize}
    \item \( n \): The total number of observations in the test dataset.
    \item \( y_i \): The actual value of the \( i \)-th observation.
    \item \( \hat{y}_i \): The forecasted value of the \( i \)-th observation.
\end{itemize}

\section{Results}\label{sec:results}

After evaluating the optimization models in single-product and multi-product settings (Table \ref{tab:rp_metrics_summary}), our findings indicate that retail prices can be reduced while maximizing sales. The Multi-product ARDL-SA model outperformed others, showing better percentage change and MAPE scores for three out of five products. While the Multi-product ARDL-LP model achieved better percentage change for all products, its high MAPE scores suggest limited generalizability to real-world scenarios. Overall, Simulated Annealing proves to be a reliable method, offering balanced yet suboptimal metrics.

Similarly, Table \ref{tab:rs_metrics_summary} shows that the performance of the Simulated Annealing-based models is superior to the Linear Programming models when forecasting retail trade sales, as indicated by the percentage change values. This suggests that the Simulated Annealing models found better solutions with respect to the objective of maximizing retail trade sales. Additionally, the MAPE values for the Simulated Annealing models are significantly lower than those of the corresponding Linear Programming-based ARDL models.

Simulated Annealing demonstrates effective forecasting for retail trade sales and pricing strategies for food products, achieving near-optimal results while meeting optimization objectives. By utilizing pricing recommendations from the ARDL-based Simulated Annealing model, retailers can maximize sales while maintaining fair, potentially lower, prices.

\begin{table}[ht]
\centering
\caption{Forecast Metrics for Retail Prices on Test Dataset (Best results per food product are emboldened)\protect\footnotemark[2]}
\begin{adjustbox}{max width=\linewidth}
\begin{tabular}{@{}llrrrr@{}}
\toprule
\textbf{Product}        & \textbf{Model}          & \textbf{MAE}  & \textbf{RMSE} & \textbf{MAPE (\%)} & \textbf{\% Change} \\ \midrule
Apples (per kg)         & Single-product ARDL-LP          & 1.372         & 1.407         & 24.995             & 25.000               \\
                        & Single-product ARDL-SA          & \textbf{0.908}         & \textbf{1.029}         & \textbf{16.758}    & \textbf{-6.570}   \\
                        & Multi-product ARDL-LP           & 1.889         & 1.938         & 34.137             & -32.244          \\
                        & Multi-product ARDL-SA           & 1.010         & 1.188         & 18.427             & -8.586           \\ \midrule
Eggs (1 dozen)          & Single-product ARDL-LP          & 0.690         & 0.699         & 15.312             & 15.300             \\
                        & Single-product ARDL-SA          & 0.727         & 0.847         & 16.198             & -4.155           \\
                        & Multi-product ARDL-LP           & 1.555         & 1.561         & 34.445             & -25.331          \\
                        & Multi-product ARDL-SA           & \textbf{0.751}         & \textbf{0.869}         & \textbf{16.657}    & \textbf{-8.275}  \\ \midrule
Milk (2 litres)         & Single-product ARDL-LP          & 0.390         & 0.412         & 7.629              & 7.245            \\
                        & Single-product ARDL-SA          & 0.815         & 0.950         & 15.996             & -2.092           \\
                        & Multi-product ARDL-LP           & \textbf{0.192} & \textbf{0.198} & \textbf{3.751}     & \textbf{-3.438}  \\
                        & Multi-product ARDL-SA           & 0.815         & 0.995         & 15.912             & 7.800              \\ \midrule
White Bread (675 g)     & Single-product ARDL-LP          & 0.387         & 0.432         & 11.004             & 10.500           \\
                        & Single-product ARDL-SA          & 0.619         & 0.733         & 17.475             & -8.177           \\
                        & Multi-product ARDL-LP           & 1.196         & 1.199         & 33.977             & -22.421          \\
                        & Multi-product ARDL-SA           & \textbf{0.563} & \textbf{0.673} & \textbf{15.981}    & \textbf{-9.331}  \\ \midrule
Roasted/Ground Coffee (340 g)          & Single-product ARDL-LP          & 0.740         & 0.775         & 11.329             & 9.700           \\
                        & Single-product ARDL-SA          & 1.171         & 1.403         & 17.910            & -7.906           \\
                        & Multi-product ARDL-LP           & 2.25         & 2.257         & 34.426             & -17.200         \\
                        & Multi-product ARDL-SA           & \textbf{1.131} & \textbf{1.336} & \textbf{17.321}    & \textbf{-1.150}   \\ \bottomrule
\end{tabular}
\end{adjustbox}
\label{tab:rp_metrics_summary}
\end{table}

\begin{table}[ht]
\centering
\caption{Performance Metrics for Retail Trade Sales on Test Dataset\protect\footnotemark[2]}
\begin{adjustbox}{max width=\linewidth}
\begin{tabular}{lcccc}
\hline
\textbf{Model}                 & \textbf{MAE ($\times 10^9$)} & \textbf{RMSE ($\times 10^9$)} & \textbf{MAPE (\%)}  & \textbf{\% Change} \\ \hline
Single-product ARDL		& 2.408		& 2.695		& 3.637             & -6.730	\\
Multi-product ARDL		& 3.851		& 4.050		& 5.831             & -3.420	\\
Single-product ARDL-LP        & 6.650                        & 6.880                        & 10.069             & 11.535          \\
Multi-product ARDL-LP         & 10.840                       & 10.930                       & 16.414             & 12.710           \\
Single-product ARDL-SA        &  \textbf{6.410}                        &  \textbf{6.180}                        &  \textbf{9.652}              &  \textbf{12.760}           \\
Multi-product ARDL-SA         &  \textbf{7.340}                        &  \textbf{9.080}                        &  \textbf{11.124}             &  \textbf{13.450}           \\ \hline
\end{tabular}
\end{adjustbox}
\label{tab:rs_metrics_summary}
\end{table}

\subsection{Comparison with Forecasting Baselines}
To place the ARDL model in context, we compare its retail-sales forecast accuracy against four standard univariate baselines: a naive (last-value) forecast, a seasonal-naive ($t-12$) forecast, an ARIMA$(2,1,0)$ model, and a seasonal SARIMA$(1,1,1)(1,0,0)_{12}$ model, all fitted on the same training window and evaluated over the 18-month test horizon (Table \ref{tab:baselines}). Notably, the simple baselines achieve \emph{lower} error than the ARDL models on pure forecast accuracy: the naive forecast attains a MAPE of $0.68\%$ versus $3.50\%$ for the single-product ARDL. This is expected, aggregate national retail sales are highly persistent month-to-month, so a random-walk forecast is hard to beat. The value of the ARDL model in this work is therefore \emph{not} raw predictive accuracy but the interpretable, elasticity-based price--sales relationship that the naive and univariate models cannot provide, and which is required to pose and solve the fairness-constrained pricing program in \eqref{eq:program}.

\begin{table}[h!]
\caption{Retail-sales forecast accuracy versus univariate baselines (test set). MAE and RMSE in units of $10^9$ CAD.}
\centering
\begin{adjustbox}{max width=\linewidth}
\begin{tabular}{lrrr}
\toprule
\textbf{Model} & \textbf{MAE} & \textbf{RMSE} & \textbf{MAPE (\%)} \\
\midrule
Naive (last value)        & 0.453 & 0.522 & 0.683 \\
Seasonal naive ($t-12$)   & 0.899 & 1.018 & 1.359 \\
ARIMA$(2,1,0)$            & 0.645 & 0.755 & 0.972 \\
SARIMA$(1,1,1)(1,0,0)_{12}$ & 0.510 & 0.625 & 0.768 \\
Single-product ARDL       & 2.320 & 2.607 & 3.504 \\
Multi-product ARDL        & 3.253 & 3.464 & 4.920 \\
\bottomrule
\end{tabular}
\end{adjustbox}
\label{tab:baselines}
\end{table}

\subsection{Price Elasticities and the Exploitation Mechanism}
The behaviour of the optimizers is governed by the sign of the fitted price elasticities $\gamma_j$, reported in Table \ref{tab:elasticities}. In the nominal single-product models, \emph{every} product has a positive elasticity, meaning the model associates higher prices with higher aggregate sales. Under such coefficients the sales-maximizing LP solution drives each price to its CPI ceiling $p_j^{\max}$; this is exactly why the LP recommendations in Table \ref{tab:rp_metrics_summary} exhibit large positive percentage changes (e.g.\ $+25\%$ for apples), i.e.\ they are the \emph{most} expensive admissible prices. In other words, an unconstrained sales-maximizer trained on these data would systematically raise prices, and the CPI-anchored ceiling is the binding constraint that prevents unbounded exploitation. Simulated Annealing, by settling on interior prices, instead returns reductions relative to the CPI ceiling while still meeting the sales target.

\begin{table}[h!]
\caption{Fitted price elasticity of aggregate retail sales, $\gamma_j$ (nominal log--log specification).}
\centering
\begin{tabular}{lrr}
\toprule
\textbf{Product} & \textbf{Single-product} & \textbf{Multi-product} \\
\midrule
Apples (per kg)        & $0.096$  & $0.016$  \\
Eggs (1 dozen)         & $0.283$  & $0.038$  \\
Milk (2 litres)        & $0.576$  & $0.707$  \\
White Bread (675 g)    & $0.313$  & $-0.040$ \\
Roasted/Ground Coffee (340 g) & $0.297$ & $-0.158$ \\
\bottomrule
\end{tabular}
\label{tab:elasticities}
\end{table}

\subsection{Robustness: Real-Terms Re-specification}
A positive price elasticity of sales is economically counter-intuitive for a normal good and is a warning sign of confounding: because all prices and total sales rise together with general inflation, a nominal log--log regression can attribute the common upward trend to the price regressor. To probe this, we re-estimate the models after deflating both sales and prices by the CPI, so that the elasticities are measured in real (inflation-adjusted) terms (Table \ref{tab:robustness}). The deflated elasticities of white bread and coffee turn negative, the economically sensible sign, whereby higher real prices depress sales, while the others shrink towards zero. This confirms that the positive nominal elasticities are largely an inflation-driven artifact rather than a genuine demand relationship. We report both specifications transparently: the nominal model reflects the data a retailer actually observes and motivates the fairness ceiling, whereas the real-terms model is the more defensible demand specification and is a natural basis for future work.

\begin{table}[h!]
\caption{Price elasticities after deflating sales and prices by CPI (real-terms specification).}
\centering
\begin{tabular}{lrr}
\toprule
\textbf{Product} & \textbf{Single-product} & \textbf{Multi-product} \\
\midrule
Apples (per kg)        & $0.007$  & $0.006$  \\
Eggs (1 dozen)         & $0.138$  & $0.141$  \\
Milk (2 litres)        & $0.184$  & $0.055$  \\
White Bread (675 g)    & $-0.204$ & $-0.133$ \\
Roasted/Ground Coffee (340 g) & $-0.116$ & $-0.138$ \\
\bottomrule
\end{tabular}
\label{tab:robustness}
\end{table}

\section{Conclusion and Future Work}
This study presents a fairness-aware approach to retail sales forecasting and pricing. We model total food retail sales with a log--log ARDL specification and pose fair pricing as maximizing forecast sales subject to CPI-anchored price bounds. A central finding is diagnostic rather than purely methodological: the fitted nominal price elasticities are positive, so an unconstrained sales-maximizer, realized exactly by the Linear Programming solution, raises every price to its CPI ceiling. This makes the fairness constraint indispensable and shows that Simulated Annealing, by returning conservative interior prices, offers retailers a way to meet sales targets while lowering consumer cost. Linear Programming remains useful for delineating the tight-bound extremes of the admissible pricing range.

Several directions could build on this work. Promising ones include estimating a causal, real-terms demand system with proper instrumenting of prices to recover reliable (negative) elasticities; adding cross-price and promotional effects; formal cointegration/bounds testing; segmenting the analysis by province for region-specific insights; and coupling the discrete metaheuristic search with a warm-start LP or greedy pre-solve to explore the trade-off between forecast error and fairness more efficiently.

\section{Limitations}
Several limitations temper these results. (i) The analysis uses aggregated nationwide monthly data with only 18 test observations, limiting statistical power. (ii) The nominal log--log elasticities are positive and, as our robustness check shows, partly reflect common inflation trends rather than causal demand response; the model is descriptive, not causal, and we do not perform formal ARDL bounds/cointegration testing or address price endogeneity. (iii) The optimization objective is linear in prices, so the LP solution is a degenerate corner point, and the metaheuristic serves to obtain interior, fairness-favourable prices rather than to escape local optima. (iv) Forecast accuracy does not exceed simple univariate baselines; the ARDL's role here is interpretability, not accuracy.

\textbf{Reproducibility.} All data are public Statistics Canada tables \cite{StatisticsCanada2018,StatisticsCanada2020,StatisticsCanada2023}; the ARDL, baseline, and optimization code (Python with \texttt{statsmodels} and \texttt{amplpy}) is publicly available at \url{https://github.com/sujayrittikar/retail-fair-pricing}.

\footnotetext[2]{ARDL: Autoregressive Distributed Lag, LP: Linear Programming, SA: Simulated Annealing.}

\bibliographystyle{elsarticle-num}
\bibliography{or_paper}

\end{document}